\documentclass{article}
\usepackage{ijcai20}
\usepackage{times}
\usepackage{soul}
\usepackage{url}
\usepackage[hidelinks]{hyperref}
\usepackage[utf8]{inputenc}
\usepackage[small]{caption}
\usepackage{graphicx}
\usepackage{amsmath}
\usepackage{amsthm}
\usepackage{booktabs}
\usepackage{algorithm}
\usepackage{algorithmic}
\usepackage{latexsym}

\usepackage{todonotes}
\usepackage{multirow}
\usepackage{enumitem}
\usepackage{newtxmath}

\DeclareMathAlphabet{\mathcal}{OMS}{cmsy}{m}{n}
\DeclareMathAlphabet{\mathbb}{U}{msb}{m}{n}
\usepackage{tabularx}
\usepackage{float}
\usepackage{adjustbox}
\usepackage{makecell}
\usepackage{subfigure}
\usepackage{microtype}

\newcommand{\vertical}[1]{\rotatebox{90}{#1}}

\title{Guided Generation of Cause and Effect}

\author{
  Zhongyang Li\textnormal{\textsuperscript{1,2}}\footnote{Performed while the first author was visiting Johns Hopkins University.}, Xiao Ding\textnormal{\textsuperscript{1}}, Ting Liu\textnormal{\textsuperscript{1}}, J. Edward Hu\textnormal{\textsuperscript{2}}{\textmd{ and }}Benjamin Van Durme\textnormal{\textsuperscript{2}}\\
  \textnormal{\textsuperscript{1}Harbin Institute of Technology, China} \\
  \textnormal{\textsuperscript{2}Johns Hopkins University, USA} \\
  \textnormal{\{zyli,xding,tliu\}@ir.hit.edu.cn, \{edward.hu,vandurme\}@jhu.edu} \\
}

\date{}
\begin{document}
\maketitle
\begin{abstract}
We present a conditional text generation framework that posits sentential expressions of possible causes and effects. This framework depends on two novel resources we develop in the course of this work: a very large-scale collection of English sentences expressing causal patterns (\textbf{CausalBank}); and a refinement over previous work on constructing large lexical causal knowledge graphs (\textbf{Cause Effect Graph}). Further, we extend prior work in lexically-constrained decoding to support \emph{disjunctive} positive constraints. Human assessment confirms that our approach gives high-quality and diverse outputs. Finally, we use CausalBank to perform continued training of an  encoder supporting a recent state-of-the-art model for causal reasoning, leading to a 3-point improvement on the COPA challenge set, with no change in model architecture.
\end{abstract}

\section{Introduction}
\begin{figure}
    \centering
    \includegraphics[width=0.8\columnwidth]{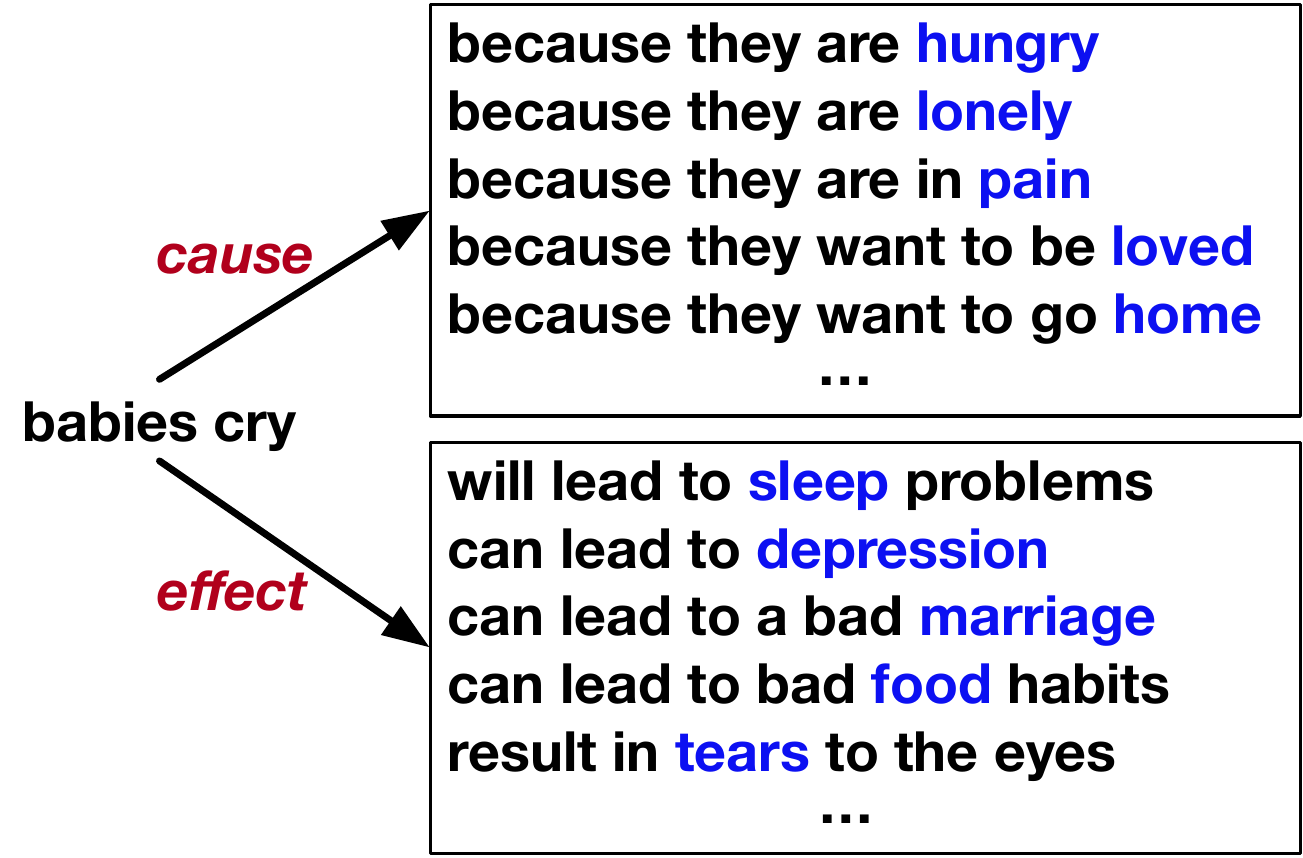}
    \vspace{-0.2cm}
    \caption{Possible causes and effects generated by our model, conditioned on the input sentence ``babies cry''. Tokens in blue are constraint keywords derived from our Cause Effect Graph, which are forced to be included in the outputs by constrained decoding.}
    \label{fig:intro}
    \vspace{-0.2cm}
\end{figure}

Causal knowledge acquisition is crucial for various Artificial Intelligence tasks, such as causal event graph construction, reading comprehension and future event prediction. We propose an approach for acquiring causal knowledge through generating multiple plausible causes (reasons, explanations) and effects (results, consequences) for a provided input sentence. As exemplified in Figure~\ref{fig:intro}, we develop two conditional decoders, one per causal direction. To train such models we mine a large-scale corpus of causal expressions from open domain web text, at a scale greatly surpassing prior work. Our goal is to generate multiple \emph{distinct} possible causes and effects, where each generated sentence is not intended to be a paraphrase of other candidates. To support this output diversity when conditioned on a single shared input sentence, we turn to lexically-constrained decoding~\cite{post2018fast,hu2019improved}, which allows for efficiently forcing a model to produce output containing one or more provided phrases. Our constraints are derived from a resource we construct for this work, replicating a prior effort in lexicalized causal knowledge graph construction~\cite{luo2016commonsense}. This graph captures causal relations as a mapping across lexical types, lemma-to-lemma, but our goal is to generate naturalistic sentences with appropriately inflected morphology: we therefore develop an approach for \emph{disjunctive} positive lexical constraints, where a decoder's output must contain one of a set of provided words or phrases. In our case, these are morphological variants of the same base lemma, but our approach should benefit other applications of lexically-constrained decoding.

While there is recent work in generating story endings conditioned on a context~\cite{guan2019story,wang19t,luo2019learning}, such work does not require generated sentences to be strictly causes or effects. The ability to propose  \emph{explanations} for an input sentence by generating multiple causes and effects complements this emerging line of research. To our knowledge,  this is the first work to consider open-ended generation of \emph{causal sentences} at a large scale. 

We evaluate through carefully designed human evaluation by comparing outputs from various baselines and our proposed model, finding that our model's outputs are preferred. We further demonstrate the usefulness of our new resource by taking a recent state-of-the-art causal reasoning system and boosting its results on the COPA test set by 3 points, relying only on continued training of the model's encoder. Our models and resources are made publicly available.\footnote{http://nlp.jhu.edu/causalbank}

\begin{figure*} 
    \centering
    \includegraphics[width=1.65\columnwidth]{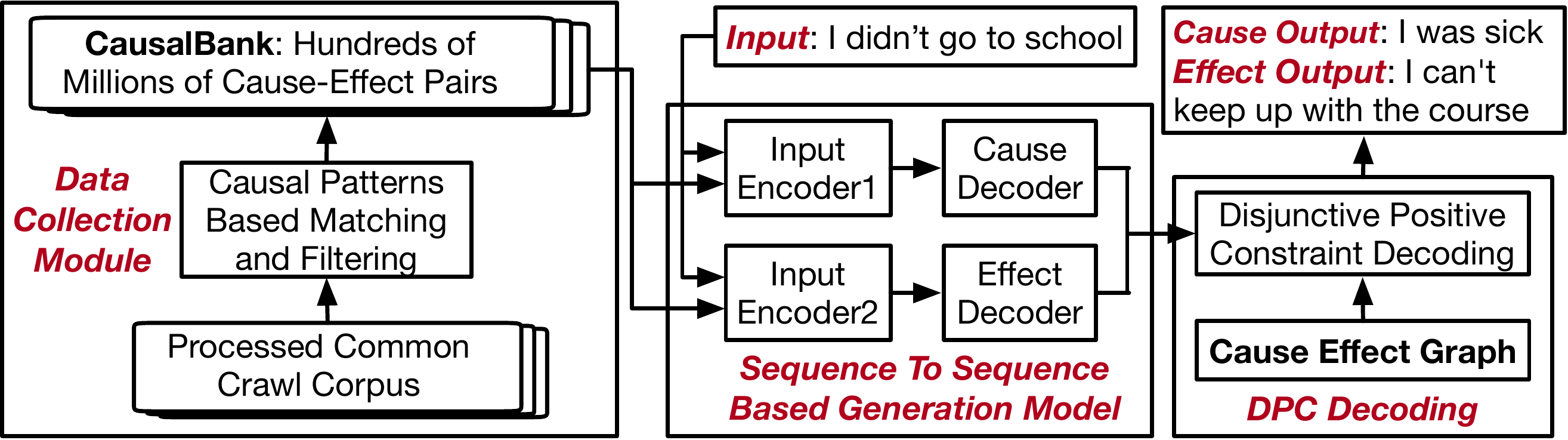}
    \vspace{-0.2cm}
    \caption{Our approach for generating plausible causes and effects.}
    \label{fig:framework}
    \vspace{-0.2cm}
\end{figure*}

In this paper, we make the following contributions:
\begin{itemize}[leftmargin=*]
\item proposing the task of open causal generation: producing possible causes and effects for any free-form textual event;
\item construction of a causal corpus (CausalBank) containing 314 million CE (cause-effect) pairs;
\item an extension to lexically-constrained decoding that supports disjunctive positive constraints (DPC);
\item human and automatic evaluations illustrating our method can generate high-quality and diverse causes and effects.
\end{itemize}

\section{Approach}
As shown in Figure \ref{fig:framework}, our proposed approach for open-ended causal generation includes a data collection module (Section 2.1), a Cause Effect Graph (Section 2.2), and two DPC (disjunctive positive constraint) decoding based Transformer encoder-decoder models (Section 2.3).

\subsection{CausalBank: A Sentential Causal Corpus}

Existing causal corpora were not built to support our goal for open-ended causal generation given any free-form textual input: as in neural machine translation (NMT), we need a large training set with millions of examples. Thus we harvest a large causal dataset from the preprocessed large-scale English Common Crawl corpus (5.14~TB) \cite{buck2014n}. The key guidelines of our dataset are as follows: 1) The causal relation is explicitly expressed in text with a causal pattern e.g. `because'; 2) The `cause' and `effect' arguments must both appear in the same sentence; 3) The `cause' and `effect' arguments can be of any length of contiguous text without overlaps between them; 4) Negative causal relations are filtered.

\begin{table}\small  
    \centering
    \begin{tabular}{c} 
         \toprule
         \textbf{Causal Pattern}\\
         \midrule
         \multirow{3}{*}{}as, as a consequence/result of, as long as, because,\\
          because of, caused by, due/owing to, in response to, \\
          on account of, result from\\
         \midrule
         \multirow{4}{*}{}accordingly, consequently, bring on/about, give rise to,\\
         induce, in order to, lead to, result in,  prevent/stop...from,\\
         and for this reason, cause, for the purpose of, if...then,\\
         ,\_so, so that, thereby, therefore, thus, hence\\
         \bottomrule 
    \end{tabular}
    \caption{Causal patterns (their morphological variants are ignored) used to get the CausalBank corpus. The first row of patterns belong to the EPC category, while the second row belong to the CPE category.}
    \label{tab:causal_pattern}
    \vspace{-0.2cm}
\end{table}

We do not rely on a supervised text extractor to pick out specific sub-spans of a sentence that represent a cause-effect pairing between propositions.\footnote{We found poor annotator agreement on span boundaries in an initial investigation on crowdsourcing data for such a system; we intend to return to this in future work, investigating improvements to our results via trained extraction models for corpus pre-processing.} We instead curate a series of patterns from previous studies \cite{mirza2014annotating,luo2016commonsense,girju2003automatic}. These patterns can be classified into two categories, according to how they are mostly used in language to convey a causal relation: 1. EPC (effect-pattern-cause) category: \textit{I am very sad} \textsc{because} \textbf{I lost my phone}; 2. CPE (cause-pattern-effect) category: \textbf{The earthquake} \textsc{resulted in} \textit{many deaths}. For EPC patterns, we simply take the text on the left of the pattern as effect, and take the text on the right of the pattern as cause. The case is reversed for CPE category patterns. These patterns (shown in Table \ref{tab:causal_pattern}) were applied to the Common Crawl corpus, followed by post-filtering: duplicate removal; filtering explicitly negated relations and verbs in passive voice; and restricting the cause and effect to each contain at least two tokens. This results in our \textbf{CausalBank} corpus, denoted here as $\mathcal{B}$, with 133 M EPC + 181 M CPE = 314 M $(c,e)$ ($c$ refers to cause and $e$ refers to effect) pairs in total. We manually evaluated 1,000 randomly sampled sentences from the corpus and found that 95\% conveyed a meaningful causal relation. 


\subsection{Cause Effect Graph: A Lexical Causal KB}

\begin{figure}
    \centering
    \includegraphics[width=0.65\columnwidth]{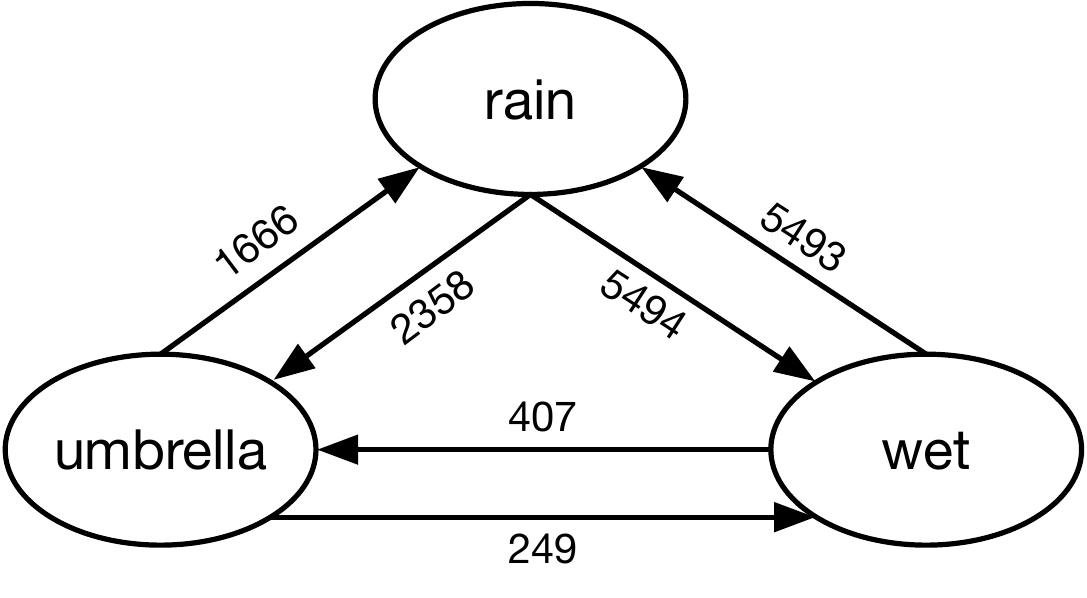}
    \vspace{-0.2cm}
    \caption{Cause Effect Graph: A lexical causal knowledge base.}
    \label{fig:causalnet}
    \vspace{-0.2cm}
\end{figure}

Following the method described in \citeauthor{luo2016commonsense}~\shortcite{luo2016commonsense} for creating a causal lexical knowledge base, we reproduce a variant of their \textbf{CausalNet} using the Common Crawl corpus \cite{buck2014n}. Given a sentence such as \emph{``The storm \textbf{caused} a tremendous amount of damage on the landing beaches.''}, this approach will  harvest the lexical pairs (\emph{storm}, \emph{tremendous}), (\emph{storm}, \emph{amount}), (\emph{storm}, \emph{damage}), (\emph{storm}, \emph{landing}), and (\emph{storm}, \emph{beach}) as causal evidence. Stop words are removed and only pairs involving nouns, verbs, adjectives and adverbs are retained. The extracted lexical pairs form a directed network of posited causal relations, where nodes in the network are lemmatized terms, and a directed edge between two terms indicates a causal relation, weighted by co-occurrence frequency. For comparison, Figure \ref{fig:causalnet} gives a similar illustration as Figure 1 in \citeauthor{luo2016commonsense}~\shortcite{luo2016commonsense}. We refer to our artifact as a \textbf{Cause Effect Graph (CEG)}; Table \ref{tab:causal_corpus} illustrates CEG contains more causal relations than CausalNet,\footnote{89.1M in contrast to 13.3M, with relations with a frequency of 5 or lower removed.} owing to the larger (5.14TB) and cleaner corpus used for extraction~\cite{buck2014n}.

\subsection{Guided Generation}


We use Sockeye \cite{hieber2017sockeye} to train  Transformer-based~\cite{NIPS2017_7181} conditional generation models, one for causes, one for effects.  Sockeye supports decoding via N-best (each step greedily chooses the top best N words in beam search based on the generated tokens) and random sampling (each step randomly sampling N words from the softmax distribution based on the generated tokens). The training data (CausalBank) is processed through Byte Pair Encoding \cite{sennrich-etal-2016-neural} to reduce vocabulary size.


\subsubsection{Disjunctive Positive Constraints Decoding} Unlike in NMT, our intended outputs for a given input are diverse in meaning: we wish to generate multiple \emph{semantically distinct} possible causes or effects. We induce diversity through hard lexical requirements during decoding, using causal keywords from our CEG as positive constraints on the output. A positive constraint forces the decoder to produce a sequence of tokens that contain the constrained sequence, which is achieved through a constrained beam search proposed by \citeauthor{post2018fast}~\shortcite{post2018fast} and made efficient by \citeauthor{hu2019improved}~\shortcite{hu2019improved}.

Unfortunately, those prior works are restricted to \emph{conjunctive} positive constraints: all items provided to the decoder \emph{must} be present in the output.  This is problematic in our case: our CEG maps lemmas to lemmas, and thus lemmas will form our constraints, but at generation time we do not require specific morphological inflections of our constrained terms.  We wish not to constrain the decoder to a particular lemma, but to allow it to choose the best morphological form as appropriate in its context. For example, when generating a cause for \emph{``I brought an umbrella''} with~\textit{rain} as the cause keyword, some valid cause sentences, e.g., \emph{``It rained''} or \emph{``It was a rainy day.''}, would not be permitted based on prior work. One may circumvent this limitation by enumerating all morphological variants of a term, then apply each in turn as a positive constraint in distinct decoding passes. However, this approach does not scale, as its run-time grows exponentially in the number of initial constraints, each with multiple morphological variants.


Here we propose a solution of \emph{disjunctive} positive constraint decoding, where each constraint is represented by a set of token sequences, and the decoder needs to include only one sequence from each set of constraints in the final output. We modify the algorithm from~\citeauthor{hu2019improved}~\shortcite{hu2019improved} to allow the decoder to explore the disjunctively constrained space in a single forward sequence, without significant computational overhead. 
In that work, constraints are represented in a trie, where each constraint is represented by a path from the root to a leaf. One or more state pointers are used to track how many tokens have been generated for each constraint, and tokens that induce more progress are prioritized in a modified beam search proposed by~\citeauthor{post2018fast}~\shortcite{post2018fast}. When a constraint is satisfied, the algorithm prunes the path representing that constraint. The distinguishing property of a disjunctive constraint is that once a sequence in a disjunctive set is satisfied, others in the set are also removed and no longer constraints. 

\begin{figure}
    \centering
    \includegraphics[width=0.99\columnwidth]{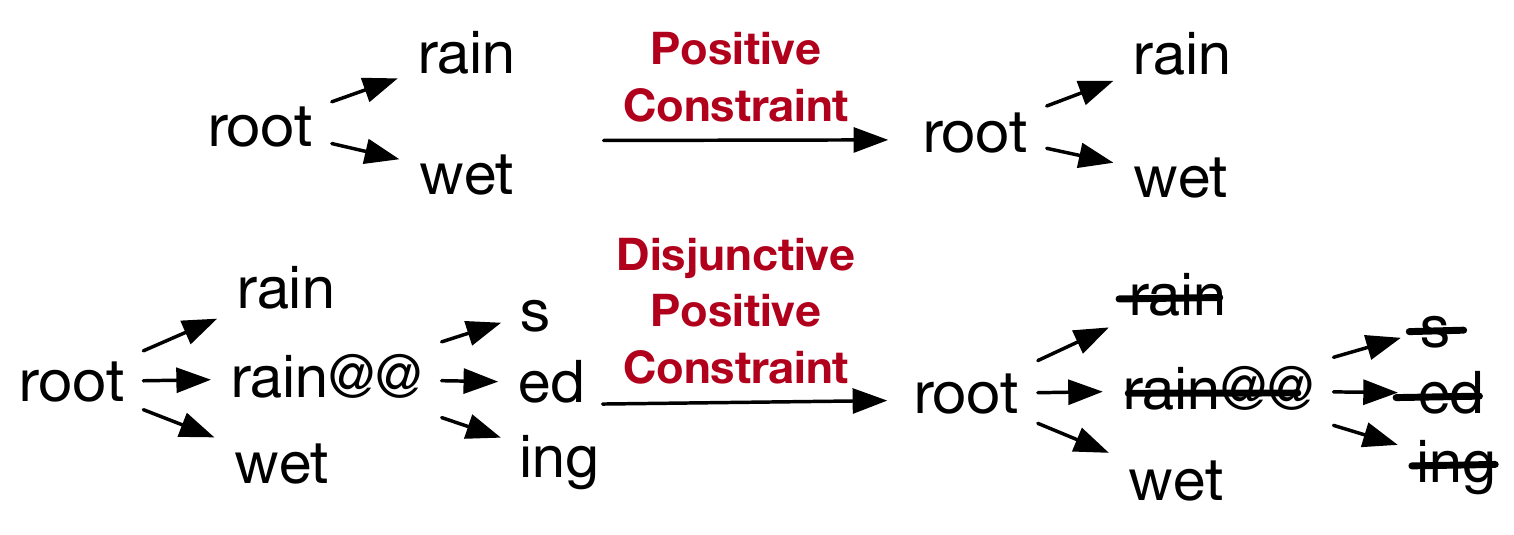}
    \vspace{-0.2cm}
    \caption{Trie states in positive constraint and disjunctive positive constraint, after generating the token `\textbf{rained}' in beam search.}
    \label{fig:disj}
    \vspace{-0.2cm}
\end{figure}

For decoding with disjunctive constraints, we represent all constrained sequences, whether they are from the same disjunctive set or not, on a single trie. When a sequence is generated, we prune \textit{all} sequences in the set as opposed to just the generated sequence. This modification gives us an efficient algorithm for applying disjunctive constraints, as illustrated in Algorithm \ref{alg:disjunctive} and Figure \ref{fig:disj}. While here we use morphological variants in our disjunctive set, our algorithm is broadly applicable for constraining on a set of synonyms or different subword segmentations of the same sequence.

\begin{algorithm}[tb]
   \caption{Decoding with Disjunctive Positive Constraints. We consider the generation of one sentence with a beam size of $1$ for simplicity. Note that while a beam size of $1$ reduces the constrained beam search, the handling of DPC is not affected.}
\label{alg:disjunctive}
\begin{algorithmic}
   \STATE {\bfseries input:} a set of disjunctive constraint sets $t$, for each set $s$ in $t$, $s_i = \{s_i^0, s_i^1, ..., s_i^n\}$ and $s_i^n=(w_i^{(n)(0)}, w_i^{(n)(1)}, ..., w_i^{(n)(m)})$ where $w_i^{(n)(m)}$ is the $m^{th}$ token in $s_i^n$, one of the sequences of the disjunctive constraint set $s_i$
   \STATE {\bfseries output:} a token sequence $o=(o_0, o_1, ..., o_k)$
   \STATE $trie \coloneqq BuildTrie(\{s_0^0, ..., s_i^n\})$
   \WHILE{$o_{k-1} !=$\texttt{EOS} and $k<k_{max}$}
   \STATE $o_k \coloneqq ConstrainedBeamSearch((o_0, ..., o_{k-1}), t)$
   \IF{$o_k$ finishes the sequence $s_p^q$}
   \FOR{$s_p^i$ in $s_p$}
   \STATE $trie \coloneqq trie.prune(s_p^i)$
   \ENDFOR
   \STATE Remove $s_p$ from $t$
   \ENDIF
   \STATE $k \coloneqq k + 1$
   \ENDWHILE
   \RETURN{$(o_0, o_1, ..., o_k)$}
\end{algorithmic}
\end{algorithm}

\paragraph{Outputs Reranking} While DPC decoding supports arbitrary number of disjunctive constraints in one beam search process, in practice only a few preferred constraints under the model will dominate any N-best output. To encourage diversity we first select a set of candidate constraint tokens from CEG, generate outputs per constraint, then merge and rerank the results. For example, if generating causes for the input sentence $i=$ \emph{``babies cry''}, we lemmatize each word in the sentence (\emph{baby} and \emph{cry}). These terms map to a set of lemmas via CEG, each associated with an observed frequency; we take the $N$-most frequent (highest weighted) such candidates: $t=\{w_1,w_2...w_N\}$. For each token $w_i$ in $t$, such as \emph{`love'}, we get a set of its morphological variants $s_i=$\{\emph{`love', `loves', `loved', `loving'}\} via the python package \texttt{patterns.en}, and pass $s_i$ as a DPC, keeping the top $M$ outputs. In total we derive $N*M$ ($N$=300 and $M$=5) sentences via $N$ beam search decodings. These sentences are ranked by their associated negative log-likelihood scores, and we return the top $K$.

\begin{table}[tp] \small 
    \centering
    \begin{tabular}{lc|cc|cc} 
        \toprule
        \multirow{2}{*}{\bf Method} & \multirow{2}{*}{\bf Dataset} & \multicolumn{2}{c}{\bf Cause} &\multicolumn{2}{c}{\bf Effect} \\
                          & &\bf Per&\bf Acc& \bf Per&\bf Acc \\
         \midrule
         RNN-LSTM & CB\_10M &66.0&29.6&55.2&32.2\\
         RNN-GRU  & CB\_10M &67.6&29.5&48.0&33.7\\
         CNN & CB\_10M &37.6&36.1&39.5&35.4\\
         Conv-Transformer  & CB\_10M &29.5&38.9&31.1&38.2\\
         Transformer & CB\_10M &\textbf{28.3}&\textbf{39.1}&\textbf{29.9}&\textbf{38.4}   \\
         \midrule
         Transformer & CB\_all &31.4&38.0&27.6&39.7 \\
         Transformer\_{BIG} & CB\_all &\textbf{29.9}&\textbf{38.5}&\textbf{26.4}&\textbf{39.8} \\
         \bottomrule 
    \end{tabular}
    \caption{Dev-set results: perplexity (\textbf{Per}), word accuracy (\textbf{Acc} (\%)).}
    \label{tab:train_dev_result}
\vspace{-0.2cm}
\end{table}

\section{CausalBERT}
Previous studies \cite{phang2018sentence,li2019story} have shown that applying intermediate auxiliary task training to an encoder such as  BERT can improve  performance on a target task. We designed an intermediate task for BERT using CausalBank $\mathcal{B}$, employing margin loss \cite{li2019learning,li2018constructing} in the objective function: $L(\Theta) = \sum_{(c,e)\in \mathcal{B}}(\max(0, m - f(c,e) + f(c',e'))) +\frac \lambda 2 ||\Theta||^2$,
where $f(c,e)$ is the score of true CE pair given by BERT model, $f(c',e')$ is the score of corrupted CE pair by replacing $c$ or $e$ with randomly sampled negative cause $c'$ or effect $e'$ from other examples in $\mathcal{B}$. $m > 0$ is the margin loss function parameter, which is set to 0.3. $\Theta$ is the set of BERT model parameters. $\lambda$ is the parameter for L2 regularization, which is set to 0.00001. 

By training BERT with this intermediate supervised task, we expect the model to acquire enhanced knowledge about the meaning of a causal relation, and can have better performance on downstream causal inference tasks.

\section{Evaluation}

We evaluate our proposed causal generation approach by both human and automatic metrics, and evaluate CausalBank by applying CausalBERT to COPA, which requires the model to choose the correct cause or effect from two candidates.

\subsubsection{Model Selection}
We first experiment on a small subset of our CausalBank corpus (CB\_10M) -- 10 million CE pairs from the causal pattern `because' -- considering different NMT encoder and decoder architectures (LSTM, CNN, Conv-Transformer~\cite{Gehring:2017:CSS:3305381.3305510}, and Transformer).\footnote{Each of these models' encoder and decoder use the same architecture, e.g. both are 6-layer LSTMs, with a hidden size and embedding size of 512. All models are trained for 10 epochs. The vocabulary size is 10,000.} For the cause generation model, $e$ is used as the source and $c$ is used as the target, which is reversed in training the effect model. Perplexity (Per) and word accuracy (Acc) are used to evaluate the model's performance. We find that Transformer constantly achieves the best performance (Table~\ref{tab:train_dev_result}).

 \begin{table}[tp] \small 
     \setlength\tabcolsep{2.0pt} 
     \centering
     \begin{tabular}{cl|cccc|cccc}
         \toprule
         \multirow{2}{*}{} & \multirow{2}{*}{\bf Method} & \multicolumn{4}{c}{\bf Cause} &\multicolumn{4}{c}{\bf Effect} \\
         & &\bf P@1&\bf P@3&\bf H &\bf Div &\bf P@1&\bf P@3&\bf H  &\bf Div \\
          \midrule
          \multirow{6}{*}{\vertical{TrainSub}}&KNN&\textbf{89.0}&67.3&\textbf{0.85}&0.11&\textbf{98.0}&71.3&\textbf{0.90}&\textbf{0.02} \\
          &GPT-2&31.0&22.3&0.39&0.13&8.0&9.3&0.30&0.11\\
          &N-Best&59.0&45.3&0.53&0.15&63.0&42.7&0.53&0.11\\
          &Random&68.0&59.3&0.66&0.11&74.0&61.7&0.70&0.09\\
          &CN-Cons&72.0&71.3&0.79&\textbf{0.02}&66.0&67.0&0.76&\textbf{0.02}\\
          &Gold-Cons&78.0&\textbf{75.3}&0.83&0.12&71.0&\textbf{73.0}&0.80&0.10\\
          \midrule
          \multirow{6}{*}{\vertical{COPA\_Dev}}&KNN&10.0&8.0&0.53&0.10&4.0&2.7&0.26&\textbf{0.01} \\
          &GPT-2&40.0&34.0&0.45&0.12&38.0&32.0&0.46&0.10\\
          &Random&66.0&53.7&0.65&0.09&62.0&46.7&0.57&0.08\\
          &N-Best&69.0&65.0&0.77&0.08&\textbf{72.0}&68.0&0.82&0.07\\
          &CN-Cons&\textbf{74.0}&70.0&0.81&\textbf{0.02}&\textbf{72.0}&\textbf{72.0}&\textbf{0.87}&0.02\\
          &Gold-Cons&73.0&\textbf{73.0}&\textbf{0.87}&0.09&\textbf{72.0}&71.3&\textbf{0.87}&0.09\\
          \bottomrule 
     \end{tabular}
     \caption{Human evaluation results of cause and effect generation.}
     \label{tab:test_result}
     \vspace{-0.2cm}
 \end{table}

Then we train two versions of Transformer on the whole CausalBank corpus (CB\_all). The small model's encoder and decoder both have 6 layers, with a hidden size and embedding size of 512. The big model's encoder and decoder have 12 layers and 4 layers, with a hidden size and embedding size of 768, leading to 134M parameters in total. The vocabulary size is 15,000. The training is stopped when the validation loss stagnates for 20,000 batches. For the cause generation model, $e$ and $c$ from only the EPC category $(c,e)$ pairs are used as the source and target. For the effect generation model, $c$ and $e$ from only the CPE category $(c,e)$ pair is used as the source and target. This setting always generates the right part of the sentence conditioned on the left part, which we find to give more reasonable outputs than the above architecture exploration experiments.   The bottom of Table~\ref{tab:train_dev_result} shows the large Transformer model constantly achieves the best performance on development set, which contains 5,000 CE pairs.

\subsubsection{Evaluating Generation}
We evaluate the large Transformer model via human assessment,  on two kinds of test sets. The first kind of test sets (TrainSub) contains 100 randomly sampled input examples from the model's training data. The second kind of test sets (COPA\_Dev) contains 100 randomly sampled examples from the development set of COPA \cite{roemmele2011choice} dataset, which are manually created gold sentences and never seen during the model's training stage. 

The compared methods include a simplified KNN method (when the input is ``babies cry'', we match sentences exactly containing the input as the retrieved neighbors, e.g. ``those babies cry loudly'', and get the corresponding causes and effects), the GPT-2 124M language model \cite{radford2019language} which can generate continuations conditioned on a start sentence (e.g. ``babies cry because''), random sampling based decoding, N-best decoding, DPC decoding with constraint tokens from CEG (CN-cons), and DPC decoding with gold answer as constraint tokens (Gold-cons).

\begin{table}[tp] \small 
    \centering
    \begin{tabular}{lc} 
         \toprule
         \textbf{Method}        &  \textbf{Acc (\%)} \\  
         \midrule
        PMI \cite{jabeen2014using} & 58.8 \\
         PMI\_EX \cite{gordon2011commonsense} & 65.4 \\
         CS \cite{luo2016commonsense}& 70.2 \\
         CS\_MWP \cite{sasaki2017handling} & 71.2 \\
         Google T5-base \cite{raffel2019exploring} & 71.2 \\
         \midrule         
         BERT-base \cite{li2019learning}& 75.4  \\
         CausalBERT-base (ours)& \textbf{78.6}  \\
         \midrule         
         Google T5-11B \cite{raffel2019exploring} & 94.8 \\
         \bottomrule 
    \end{tabular}
    \caption{Results on COPA-Test, contrasting prior results to a model by Li \emph{et al.} built atop BERT-base. This model is improved by 3 points through adoption of CausalBERT.}
    \label{tab:copa_results}
    \vspace{-0.2cm}
\end{table}

Four graduate students from the NLP field were used in annotation. Each was asked to give a score from $\{0,1,2\}$ for the generated \{input, cause/effect\} pair, where the guidelines are (take cause generation for example): if the generated answer does not make sense or can never be a reasonable cause, reason or explanation for the input event, give a score of 0; if the generated answer has grammatical errors but can be a reasonable cause, reason or explanation for the input event under some rare conditions (or beyond commonsense), give a score of 1; if the generated answer is a fluent sentence and can be a reasonable cause, reason or explanation with high probability, give a score of 2. Each pair was labeled by two annotators, and we average the judgments over two annotators per pair. The cohen's kappa score is 0.53.

Table \ref{tab:test_result} shows the human evaluation results. Three metrics are adopted: Precision at 1 \textbf{P@1} (an average score of 1.5 or above is seen as a valid causal answer); \textbf{P@3}; and average human score for each evaluated pair ({\bf H}). For the TrainSub test set, the KNN method shows strong performance, especially for P@1 and the human scores. However, KNN performs worse for P@3, due to the absence of many possible answers for the same input. Meanwhile, our two versions of DPC decoding strategies (CN-cons, Gold-Cons) also show relatively better performance compared to other generation methods (GPT-2, Random and N-best decoding). KNN performs poorly on the COPA dev set, because most of the inputs never appear in the training data. However, CN-Cons and Gold-Cons can still achieve good performance.

\paragraph{Lexical Diversity} 
We used a modified BLEU score to evaluate lexical diversity (\textbf{Div} in Table \ref{tab:test_result}) where a lower score means a greater lexical diversity. Specifically, we calculate the associated BLEU-1 score between the gold answers and the generated top 3 outputs without brevity penalty. This modification ensures that we don't reward shorter outputs. In most cases, CN-Cons gets the lowest \textbf{Div} scores, showing that our DPC decoding and constraint tokens from CEG together, allows us to explore more in the causes and effects space, and generate more diverse outputs. Also we find that all of these BLEU scores are very low, compared with the BLEU scores in previous text generation studies \cite{hu2019parabank,NIPS2017_7181}. This is because our generation task is open-ended (as illustrated in Figure \ref{fig:intro}). 

\paragraph{Evaluating CausalBank} 
Table \ref{tab:copa_results} shows our CausalBERT results on COPA test. Compared with prior strong knowledge-driven baseline methods, a BERT-base model trained with a margin-based loss \cite{li2019learning} achieved good performance. Following the experimental settings of \citeauthor{li2019learning}~\shortcite{li2019learning}, when training the BERT-base model with additional CE pairs from CausalBank, we get an improvement of 3.2\%, from 75.4\% to 78.6\%, showing that our corpus successfully augments BERT base to make it better for causal inference, which is a sign the corpus contains useful causal knowledge. We find that the number of CE pairs in the intermediate task matters: performance first improves and then decreases, with more training data added. \footnote{This was not observed in related studies \cite{phang2018sentence,li2019story}, where all training examples from the Multi-NLI dataset were used as an intermediate task. Similar behavior was observed in NMT in continued training for domain adaptation \cite{thompson-etal-2019-overcoming}. We believe ours to be a similar setting, where the ``in-domain'' causal data overwhelms the benefits of pretraining; adapting strategies from Thompson \emph{et al.} is an avenue for future work.}
We get the best performance of 78.6\% with 40 K training CE pairs. Though our result still has a big gap from the current SOTA performance on COPA (94.8\% from the largest google T5-11B model), the intent of our experiment is just to illustrate how the only difference was in altering the pre-training with CausalBank. One could possibly get a SOTA model based on our corpus and the google T5 model, if publicly available.

\section{Related Work}
\paragraph{Conditional Text Generation}
Such efforts cover a large body of work, including machine translation, 
response generation  
and paraphrase generation.  
Most related is conditional story generation \cite{guan2019story,wang19t,luo2019learning,li2018generating}, which aims to generate story continuations based on a given context. These works do not require generated sentences to be strictly causes or effects. 

For causal generation, \citeauthor{rashkin2018event2mind}~\shortcite{rashkin2018event2mind} aimed to generate the likely intents and reactions of the event's participants, given a short free-form textual event. \citeauthor{sap2019atomic}~\shortcite{sap2019atomic} trained a multi-task model for fine-grained kinds of \textit{If-Then} commonsense reasoning. However, the causal semantics considered in their work are restricted to a narrow space, and their models are trained on no more than one million examples. Further, their resource was based-on crowdsourcing, which carries risks of human bias~\cite{rudinger-etal-2017-social,poliak-etal-2018-hypothesis}. We harvest a significantly larger, open coverage causal corpus,\footnote{While we avoid pitfalls of elicitation, we acknowledge that like any corpus-extracted resource ours may suffer from \emph{reporting bias} \cite{Gordon2013Reporting}: some types of causes or effects that are known to humans but rarely or ever explicitly stated.} related in approach to DisSent~\cite{nie-etal-2019-dissent} but larger, focused on causality, and aimed primarily at generation rather than sentence representation learning.

Of various efforts in guided generation \cite{ammanabrolu2019guided,tang-etal-2019-target,clark2018neural,hu2019parabank}, lexically-constrained decoding \cite{hokamp2017lexically} is a modification of beam search originating in neural machine translation which allows the user to specify tokens that must (or must not) appear in the decoder's output. 

\citeauthor{post2018fast}~\shortcite{post2018fast} proposed a variant of lexically-constrained decoding that reduced complexity from linear to constant-time, which was made more efficient by \citeauthor{hu2019improved}~\shortcite{hu2019improved}. 
We introduce an extension to lexically-constrained decoding that supports disjunctive positive constraints for multiple optional constraint keywords.

\begin{table}[tp]\small  
    \centering
    \begin{tabular}{lr} 
         \toprule
         \textbf{Sentential Causal Resource} & \textbf{\# CE Pairs}\\
         \midrule
         TCR \cite{ning2018joint} &172 \\
         SemEval-2007 Task4 \cite{girju2007semeval}&220\\
         Causal-TimeBank \cite{mirza2014annotating}&318\\
         CaTeRS \cite{mostafazadeh2016caters}&488\\
        EventCausalityData \cite{do2011minimally}&580\\
        RED \cite{o2016richer}&1,147\\
         SemEval2010 Task8 \cite{hendrickx2009semeval}&1,331\\ 
         BECauSE 2.0 \cite{dunietz2017because}&1,803\\
        EventStoryLine \cite{caselli2017event}&5,519\\
        PDTB 2.0 \cite{prasad2008penn}&8,042\\        
         Altlex \cite{hidey2016identifying}&9,190\\
         PDTB 3.0 \cite{webber2019penn}&13 K\\
         DisSent \cite{nie-etal-2019-dissent}&167 K\\
         \textbf{CausalBank} (Ours)&\textbf{314 M}\\
          \midrule
          \textbf{Causal Knowledge Graph} & \textbf{\# CE Edges}\\
         \midrule
          Event2mind \cite{rashkin2018event2mind}&25 K\\
         ConceptNet 5.7 \cite{speer2017conceptnet}&473 K\\
         ASER Core \cite{zhang2019aser}&494 K\\
         Atomic \cite{sap2019atomic}&877 K\\
         CausalNet \cite{luo2016commonsense}&13.3 M\\
         \textbf{Cause Effect Graph} (Ours)& \textbf{89.1 M}\\
         \bottomrule 
    \end{tabular}
    \caption{Contrasting size with example prior works: only the causal portion of these corpora are listed. The top are sentential causal corpora, while the bottom are graph-structure causal knowledge bases.}
    \label{tab:causal_corpus}
    \vspace{-0.2cm}
\end{table}

\paragraph{Sentential Causal Resources}
Existing causal corpora differ in their annotation guidelines and how they are constructed: (1) whether they consider only explicit or also implicit causal relations; (2) whether they consider only intra-sentence relations or if relations can cross sentences; (3) whether the annotation unit is word level or sentence level; and (4) whether the corpus is constructed automatically or by human effort. Ours is concerned with explicit only relations, within a single sentence, relating one part of a sentence to another, and employs constructed patterns but not sentence-level human annotation.

Already mentioned are recent crowdsourcing efforts \cite{rashkin2018event2mind,sap2019atomic}. More related are PDTB \cite{prasad2008penn} and BECauSE \cite{dunietz2017because}, but where our resource goal is a much larger corpus, for the purpose of training a neural text generation model. Most related would be the extractive approach of DisSent \cite{nie-etal-2019-dissent}, but where we focus specifically on causality, and derive a much larger corpus. \cite{bethard2008learning} tagged a small corpus of event pairs conjoined with ``and'' as causal or not causal. CaTeRS \cite{mostafazadeh2016caters} included causal relations from a commonsense reasoning standpoint. Richer Event Description \cite{o2016richer} integrates real-world temporal and causal relations between events into a unified framework. Table~\ref{tab:causal_corpus} contrasts the size of causal portion of prior resources with our own.

\paragraph{Lexical Causal Resources}
Lexical semantic resources may encode causal properties on verbs (e.g., \cite{schuler2005verbnet,bonial2014propbank}) and prepositions (e.g., \cite{schneider2015hierarchy}). Force dynamics theory \cite{talmy1988force} from cognitive psychology posits three primary kinds of causal semantics \cite{wolff2007representing} -- CAUSE, ENABLE and PREVENT -- which were lexicalized as causal verbs \cite{wolff2003models}. 
The annotation scheme of \citeauthor{dunietz2017because}~\shortcite{dunietz2017because} distinguishes three types of causal semantics: CONSEQUENCE, MOTIVATION, and PURPOSE. In PDTB 2.0 \cite{prasad2008penn}, ``CONTINGENCY'' has two subtypes (``Cause'' and ``Condition''). FrameNet \cite{baker2014framenet} represents causal relations through a variety of unrelated frames (e.g., CAUSATION and THWARTING) and frame roles (e.g., PURPOSE and EXPLANATION). These efforts motivate our own causal patterns, categorized into: \textbf{CAUSE} (e.g. cause, result in, lead to), \textbf{EXPLANATION} (e.g. because, due to), \textbf{CONDITION} (e.g. if-then, as long as), \textbf{PURPOSE} (e.g. in order to, for the purpose of), and \textbf{PREVENTION} (e.g. stop/prevent-from).

\paragraph{Causal Knowledge Acquisition}
Causal knowledge acquisition \cite{radinsky2012causality,radinsky2013mining} 
is crucial for many AI systems, 
and it is often acquired via text. 
%
%
\citeauthor{hashimoto2014toward}~\shortcite{hashimoto2014toward} and \citeauthor{kruengkrai2017improving}~\shortcite{kruengkrai2017improving} applied supervised learning techniques using a benchmark training data with over 100K human-annotated CE pairs. \citeauthor{dasgupta2018automatic}~\shortcite{dasgupta2018automatic} explored general causal extraction using 5,000 labelled sentences. \citeauthor{do2011minimally}~\shortcite{do2011minimally} is an example of a minimally supervised approach. Recent studies \cite{dunietz2017automatically,dunietz2018deepcx} explored new supervised approaches on the BECauSE 2.0 \cite{dunietz2017because} corpus.

\citeauthor{church1990word} ~\shortcite{church1990word} proposed the use of pointwise mutual information (PMI) for mining patterns via text co-occurrence. Many works have followed this strategy, e.g. ~\cite{chambers2008unsupervised,riaz2010another,gordon2011commonsense,do2011minimally,luo2016commonsense}. Others have mined patterns via discourse patterns in the form of `A led to B', `if A then B', etc., e.g., \cite{khoo2000extracting,girju2003automatic,zhao2017constructing}). 
See \citeauthor{asghar2016automatic}~\shortcite{asghar2016automatic} for review. %
Such efforts relate most closely to our CEGraph component, rather than our overall framework.  Our concern is the generation of diverse potential causes and effects as natural language statemnts.


\section{Conclusion}
We investigate open causal generation for free-form textual input, and build a large sentential causal corpus which we used to train a generative model. We introduced a novel extension to lexically-constrained decoding that supports disjunctive positive constraints, where generated output is forced to contain one of a set of candidates. Automatic and human evaluations show that our method can generate high-quality and diverse causes and effects for new inputs.

\section*{Acknowledgements}
We acknowledge the support of the National Key Research and Development Program of China (SQ2018AAA0101901), the National Natural Science Foundation of China (NSFC) via Grant 61976073 and 61702137; the China Scholarship Council; and DARPA KAIROS (Hu and Van Durme).
\bibliographystyle{named}
\bibliography{ijcai20}
\end{document}